\pdfoutput=1

\documentclass[11pt]{article}

\usepackage[preprint]{acl}

\usepackage{times}
\usepackage{latexsym}
\usepackage{hyperref}
\usepackage{graphicx}
\usepackage{float}

\usepackage{algorithm}
\usepackage{algpseudocode}
\usepackage{amsmath}
\usepackage{xcolor} 

\usepackage[T1]{fontenc}
\usepackage{natbib} 

\usepackage[utf8]{inputenc}

\usepackage{microtype}

\usepackage{inconsolata}

\usepackage{graphicx}

\usepackage{tikz}
\usepackage{varwidth}
\usepackage{lipsum}
\usepackage{geometry}
\usepackage{placeins} 

\title{SALSA: Single-pass Autoregressive LLM Structured Classification}

\author{
  Ruslan Berdichevsky \\
  Dream Security Ltd. \\
  Tel Aviv, Israel \\
  \texttt{ruslan@dreamgroup.com}
  \And
  Shai Nahum-Gefen \\
  Dream Security Ltd. \\
  Tel Aviv, Israel \\
  \texttt{shai@dreamgroup.com}
  \And
  Elad Ben Zaken \\
  Dream Security Ltd. \\
  Tel Aviv, Israel \\
  \texttt{elad@dreamgroup.com}
}

\begin{document}
\maketitle

\begin{abstract}

Despite their impressive generalization capabilities, instruction-tuned Large Language Models often underperform on text classification benchmarks. We introduce SALSA, a coherent pipeline that combines structured prompting, class-to-token mapping, and parameter-efficient fine-tuning, thereby avoiding cold-start training. Each class label is mapped to a distinct output token, and prompts are constructed to elicit a single-token response. During inference, the model’s output is projected only onto the logits of the relevant class tokens, enabling efficient and accurate classification in a single forward pass. SALSA achieves state-of-the-art results across diverse benchmarks, demonstrating its robustness and scalability for LLM-based classification applications.

\end{abstract}

\section{Introduction}
Text classification is fundamental in natural language processing (NLP), underpinning applications such as spam detection, sentiment analysis, dialogue safety, and content moderation. Traditional methods involving handcrafted rules and features were limited by scalability and labor intensity. The emergence of deep learning transformed the field by enabling automated feature extraction through models such as word2vec \citep{DBLP:journals/corr/abs-1301-3781}, ELMo \citep{peters-etal-2018-deep}, and transformer-based architectures such as BERT \citep{devlin-etal-2019-bert} and GPT \citep{brown2020language}, which deliver exceptional performance.

With the advent of Large Language Models (LLMs), particularly open-ended generative models, the capabilities of NLP systems have expanded significantly. These models, pre-trained on extensive corpora, encapsulate a wealth of transferable knowledge that can be leveraged for diverse downstream tasks, including text classification. Despite this, the effective adaptation of open-ended generative LLMs for classification still poses challenges, requiring efficient input representation and fine-tuning strategies.

Recent methods commonly utilize chain-of-thought (CoT) prompting \citep{wei2022chain}, effective for reasoning, but computationally inefficient for classification. Such approaches also neglect valuable information in output distribution. In contrast, discriminative approaches (e.g., \citealp{pawar-etal-2024-generate}) remain underexplored.

In this paper, we introduce SALSA (Single-pass Autoregressive LLM Structured Classification), a method that adapts instruction-tuned, decoder-only LLMs for text classification. SALSA integrates structured prompt construction, targeted logit analysis, and fine-tuning into a unified pipeline. Its prompt-driven design enables strong zero-shot performance, providing a favorable initialization for subsequent tuning. Though built for generation, decoder-only LLMs can act as effective classifiers—matching or exceeding state-of-the-art results across benchmarks. By relying on a single forward pass, SALSA also offers a more efficient alternative to generation-based methods.

\section{Background}

Early NLP approaches used handcrafted features, deep learning then introduced RNNs and CNNs, improving classification \citep{kim-2014-convolutional}. Transformer-based models, introduced by Vaswani et al. \citep{NIPS2017_3f5ee243}, revolutionized NLP by utilizing self-attention mechanisms for contextualized embeddings. Models like BERT represented a major leap forward by introducing bidirectional context understanding through unsupervised pretraining on large-scale corpora. Autoregressive transformer models like XLNet \citep{yang2019xlnet} demonstrated the benefits of autoregressive pretraining, outperforming traditional methods in classification tasks.

It has since been shown that large language models implicitly encode world knowledge, which can be extracted via their output logits \citep{petroni2019language}. Reformulating cloze-style tasks as multiple-choice classification has proven effective \citep{robinson2022leveraging}, but the reliability of such approaches is highly sensitive to prompt structure \citep{cao-etal-2021-knowledgeable}.

Instruction tuning was a key breakthrough in demonstrating that language models can generalize across tasks when aligned with task-specific instructions \citep{wei2021finetuned}. Building on this insight, decoder-only LLMs such as GPT \citep{brown2020language}, LLaMA \citep{touvron2023llamaopenefficientfoundation, grattafiori2024llama3herdmodels}, and Gemma \citep{gemmateam2024gemmaopenmodelsbased} have redefined the field, supporting zero-shot and in-context learning with strong generalization capabilities across a wide range of NLP tasks.

Parameter-efficient methods like BitFit \citep{ben-zaken-etal-2022-bitfit} and Low-Rank Adaptation, LoRA \citep{hu2021lora}, further limit overfitting by reducing the number of trainable parameters, ensuring stable fine-tuning especially in low-data scenarios. They also enable cost-effective deployment across tasks, requiring only minimal parameter swaps while leaving the base model intact.

When comparing results, fine-tuned encoder-based LLMs have achieved better performance in classification tasks, such as in the GLUE benchmark \citep{glue}. Surprisingly, the much larger instruction decoder-only LLMs, which often outperform encoder-based LLMs in several tasks, generally fail to achieve competitive classification results \citep{bucher2024fine}.

Our work aims to bridge the gap between the potential of instruction-decoder-only LLMs and the performance of classification tasks, both in terms of quality and efficiency.

\section{Method}
\begin{figure*}[h!]
  \centering
  \includegraphics[width=\textwidth]{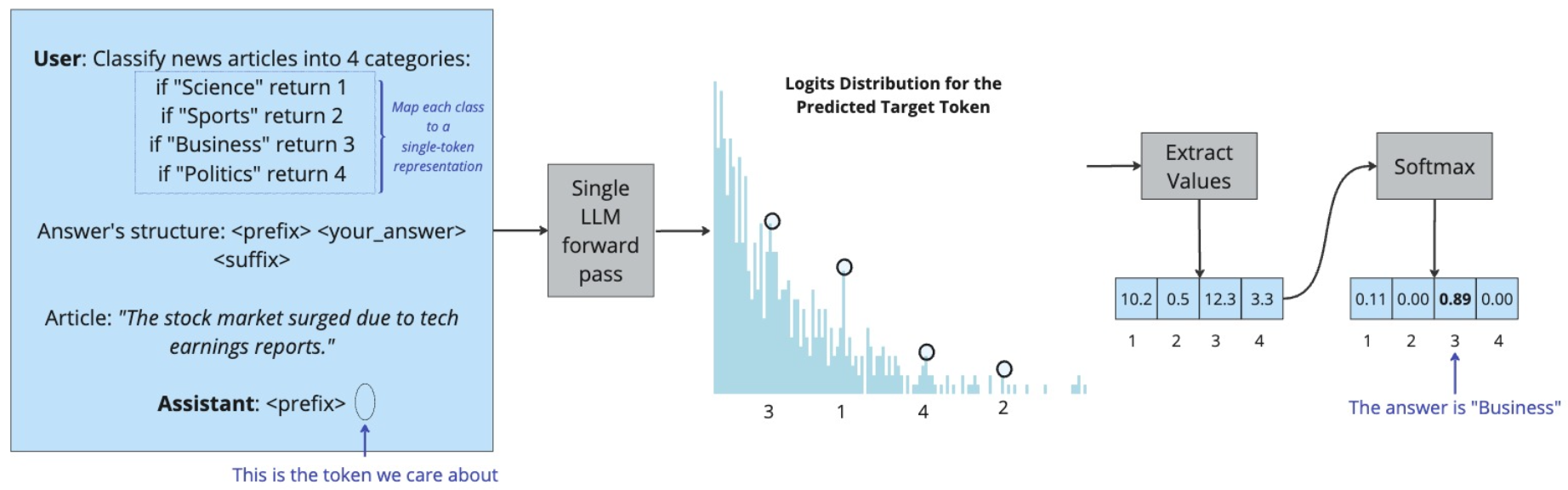}
  \caption{SALSA single-token classification pipeline: each category is mapped to a distinct token, and the LLM’s logits determine the predicted label in one forward pass.}
  \label{fig:single_class_method_diagram}
\end{figure*}

SALSA leverages the internal knowledge of LLMs by using their output estimated distribution to perform classification in a single forward pass per query. Our method employs LoRA for efficient parameter updates and knowledge exposure, allowing SALSA to deliver competitive performance.

\noindent\textbf{Prompt Construction.}\quad
We design a structured instruction prompt that encapsulates the task. The prompt first provides a clear task description, then maps each class to a unique single-token representation, and finally specifies the expected answer format, including fixed prefix and suffix elements. A structured response containing a placeholder token is appended to complete the prompt. This process is illustrated in Figure~\ref{fig:single_class_method_diagram}. A detailed example is provided in ~\ref{prompt_exm}

\noindent\textbf{Forward Pass, Filtering, and Classification.}\quad
We perform a single forward pass through the LLM to extract the logits for the placeholder token, which represent the model's predictions. These logits are then filtered based on the prompt's mapping and normalized via softmax to yield an estimated probability distribution over the classes. The final prediction corresponds to the class with the highest probability.

\noindent\textbf{Training.}\quad
We optimize our model using a backpropagation-based procedure (see \ref{train_alg}). In particular, we employ LoRA in conjunction with a cross-entropy loss function. The loss is defined as follows:
\begin{equation}
 \label{eq:crossentropy_loss}
 \mathcal{L} = -\frac{1}{N} \sum_{i=1}^N \sum_{c=1}^C y_{i,c} \log(\hat{P}_{i,c})
\end{equation}
where \( N \) is the number of samples, \( C \) is the number of classes, \( y_{i,c} \) represents the ground truth labels, and \( \hat{P}_{i,c} \) denotes the predicted probabilities. See ~\ref{training_details} for more details.

\noindent\textbf{Controlling the Precision–Recall Trade-off.}\quad
Adjusting decision threshold values offers precise control over the trade-off between precision and recall. This flexibility allows the model to be tailored to specific application needs, enabling tuning to optimize performance based on the desired balance.

\noindent\textbf{Efficient Single-Pass Inference.}\quad
SALSA eliminates autoregressive overhead by computing all logits in a single forward pass, reducing latency, resource use and cost. Mapping classification to a single-token output ensures only valid class tokens are considered, enhancing efficiency and correctness.

\section{Experiments and Results}
\subsection{Datasets}
We evaluated SALSA on multiple text classification datasets, including a subset of GLUE \citep{glue}, covering SST-2 \citep{sst2}, MRPC \citep{mrpc}, QQP \citep{qqp}, MNLI \citep{mnli}, QNLI \citep{qnli}, and RTE \citep{rte}. Additional datasets included AG's News \citep{yelp} for topic classification, IMDb \citep{maas-EtAl:2011:ACL-HLT2011} for binary sentiment analysis, and Yelp-5 \citep{yelp} for multi-class sentiment analysis. We further included MedNLI \citep{romanov-shivade-2018-lessons} for clinical natural language inference, MedMCQA \citep{pal2022medmcqalargescalemultisubject} for multiple-choice medical question answering, and HateXplain \citep{DBLP:journals/corr/abs-2012-10289} for hate speech and offensive language detection. For more details see section \ref{datasets_section}.

\subsection{Analysis}
\begin{table*}[htbp!]
  \centering
  \scalebox{0.89}{\begin{tabular}{clccccccc}
    \hline
    \hline
     & & \textbf{QQP} & \textbf{SST-2} & \textbf{RTE} & \textbf{MRPC} &\textbf{QNLI} & \textbf{MNLI\textsubscript{M}} & \textbf{MNLI\textsubscript{MM}} \\
    \hline
    \hline
    (V) & Zero Shot & 81.4 & 94.9 & 86.3 & 77.0 & 90.7 & 81.9 & 80.9 \\
    (V) & Few Shot & 81.5 &  96.1 & 85.2 & 77.2 & 91.4 & 80.1 & 80.2 \\
    (V) & RoBERTa\textsubscript{LARGE} & 92.2 & 96.4 & 86.6 & 90.9 & 94.7 & 90.2 & 90.2 \\
    (V) & ALBERT & 92.2 & 96.9 & 89.2 & 90.9 & 95.3 & 90.8 & 90.8 \\
    (V) & XLNet & 92.3 & 97.0 & 85.9 & 90.8 & 94.9 & 90.8 & 90.8 \\
    \hline
    (V) & SALSA Zero Shot $^{\dagger}$ & 82.1 & 95.0 & 90.6 & 76.4 &  92.7 &  84.1 &  83.1  \\   
    (V) & SALSA Few Shot $^{\dagger}$ & 83.3 & 95.4 & 92.0 & 80.1 &  92.9 &  86.7 &  86.3  \\  
    (V) & SALSA & \textbf{92.4$\pm$0.2} & \textbf{97.1$\pm$0.2} & \textbf{94.2$\pm$0.4} & \textbf{91.7$\pm$0.5} & \textbf{96.7$\pm$0.2} & \textbf{92.8$\pm$0.3} & \textbf{92.6$\pm$0.2} \\
    \hline
    \hline
    (T) & BERT\textsubscript{LARGE} & 89.3 & 94.9 & 70.1 & 85.4 & 92.7 & 86.7 & 85.9 \\
    (T) & T5-11B & 90.6 &  97.5 & 92.8 & 90.4 & 96.9 & 92.2 & 91.9 \\
    (T) & Turing ULR v6 & 90.9 & 97.5 & 93.6 & 92.3 & 96.7 & 92.5 & 92.1 \\
    (T) & Vega v1 & \textbf{91.1} & \textbf{97.9} & 92.4 & \textbf{92.6} & 96.7 & 92.2 & 91.9 \\
    (T) & Turing ULR v5 & 91.1 & 97.6 & 94.1 & 91.7 & \textbf{97.9} & 92.6 & \textbf{92.4} \\
    \hline
    (T) & SALSA & 90.9 & \textbf{97.9} & \textbf{94.8 }& 91.1 & 97.1 & \textbf{92.7} & 92.0 \\
    \hline
    \hline
    
  \end{tabular}}
  \caption{Performance metrics of SALSA compared to baseline models across multiple GLUE Benchmark datasets. Results are reported separately for the validation (V) and test (T) sets, with accuracy as the key evaluation metric. SALSA achieves state-of-the-art performance on all validation tasks and outperforms competitors on 3 out of 7 test tasks. Test set results are benchmarked against the top 3 GLUE leaderboard models as of January 27, 2025. $^{\dagger}$No fine-tuning applied.}

  \label{tab:comparison_to_other_methods}
\end{table*}
\begin{table}[h]
  \centering
  \scalebox{0.97}{\begin{tabular}{lccc}
    \hline
    \hline
      & \textbf{AG News} & \textbf{IMDb} & \textbf{Yelp-5} \\
    \hline
    \hline
    Zero Shot & 88.8  & 95.2 & 62.7 \\
    XLNet & 95.5 & 96.8 & 72.9 \\
    \hline
    SALSA & \textbf{95.9$\pm$0.1} & \textbf{97.6$\pm$0.1} & \textbf{74.2
     $\pm$0.2} \\

    \hline
    \hline
  \end{tabular}}
  \caption{Accuracy on AGNews, IMDb, and Yelp-5 test datasets.}
  \label{tab:other_benchmarks}
\end{table}

\begin{table}[h]
  \centering
  \scalebox{0.87}{
  \begin{tabular}{lccc}
    \hline
    \hline
     & \textbf{MedNLI} & \textbf{MedMCQA} & \textbf{HateXplain} \\
    \hline
    \hline
    Zero-Shot & 83.4 & 70.3 & 51.5 \\
    SOTA & 90.2$^{\dagger}$ & 73.6$^{\dagger}$ & 70.4$^{\dagger}$ \\
    \hline
    SALSA & \textbf{91.3$\pm$0.4} & \textbf{74.1$\pm$0.3} & \textbf{71.8$\pm$0.4} \\
    \hline
    \hline
  \end{tabular}}
  \caption{Accuracy on MedNLI, MedMCQA, and HateXplain test datasets.
$^{\dagger}$SOTA sources: GatorTron-large for MedNLI~\citep{yang2022gatortronlargeclinicallanguage}, GPT-4 for MedMCQA~\citep{nori2023capabilitiesgpt4medicalchallenge}, and BERT-MRP for HateXplain~\citep{kim2022hatespeechmaskedrationale}.}
  \label{tab:other_benchmarks_fields}
\end{table}

In this section, we delve into a comprehensive analysis of SALSA by examining performance metrics, convergence efficiency, and other key aspects across various benchmarks.

\noindent\textbf{State-of-the-Art Results.}\quad
SALSA demonstrates state-of-the-art performance across multiple text classification benchmarks, as outlined in Table \ref{tab:comparison_to_other_methods} (and Table \ref{tab:other_benchmarks}).

The method consistently outperforms existing models, including T5-11B \citep{t5}, XLNet \citep{yang2019xlnet}, RoBERTa\textsubscript{LARGE} \citep{liu2019roberta}, and ALBERT \citep{DBLP:journals/corr/abs-1909-11942}. Furthermore, we compared SALSA against the top three performers on the GLUE benchmark, Turing ULR v6 \citep{TULRv6}, Vega v1 \citep{vega_v1}, and Turing ULR v5 \citep{TULRv5}, and SALSA outperforms them all in 3 of 7 tasks. 

Furthermore, we evaluated SALSA on three domain-specific benchmarks: MedNLI, MedMCQA, and HateXplain. As shown in Table~\ref{tab:other_benchmarks_fields}, SALSA outperforms previous SOTA, demonstrating strong generalization in diverse NLP tasks.

For each validation set experiment, we train the model five times with different random seeds and report the average performance on the validation set. For test set experiments, we evaluate the model that achieves the highest results on the validation set using the GLUE test set evaluation server. These findings validate the efficiency and robustness of SALSA in leveraging generative LLMs for classification tasks.

\noindent\textbf{Zero-Shot and Few-Shot Classification.}\quad
To further evaluate SALSA, we conducted zero-shot and few-shot classification experiments using Meta’s Instruct LLaMA 3.3 70B model.

In the zero-shot setting, the model received structured prompts containing task instructions and class options, without labeled examples. For few-shot classification, we added ten balanced random examples as in-context demonstrations. In both cases, the model generated open-ended responses that we parsed to extract the predicted classes.

We also applied SALSA in both settings, without fine-tuning. The zero-shot variant included only task instructions and class labels; the few-shot variant appended a few formatted examples. As shown in Table~\ref{tab:comparison_to_other_methods}, SALSA clearly outperforms standard prompting approaches.

\begin{figure}[ht]
    \centering
     \hspace*{-0.43cm}
    \includegraphics[width=0.51\textwidth]{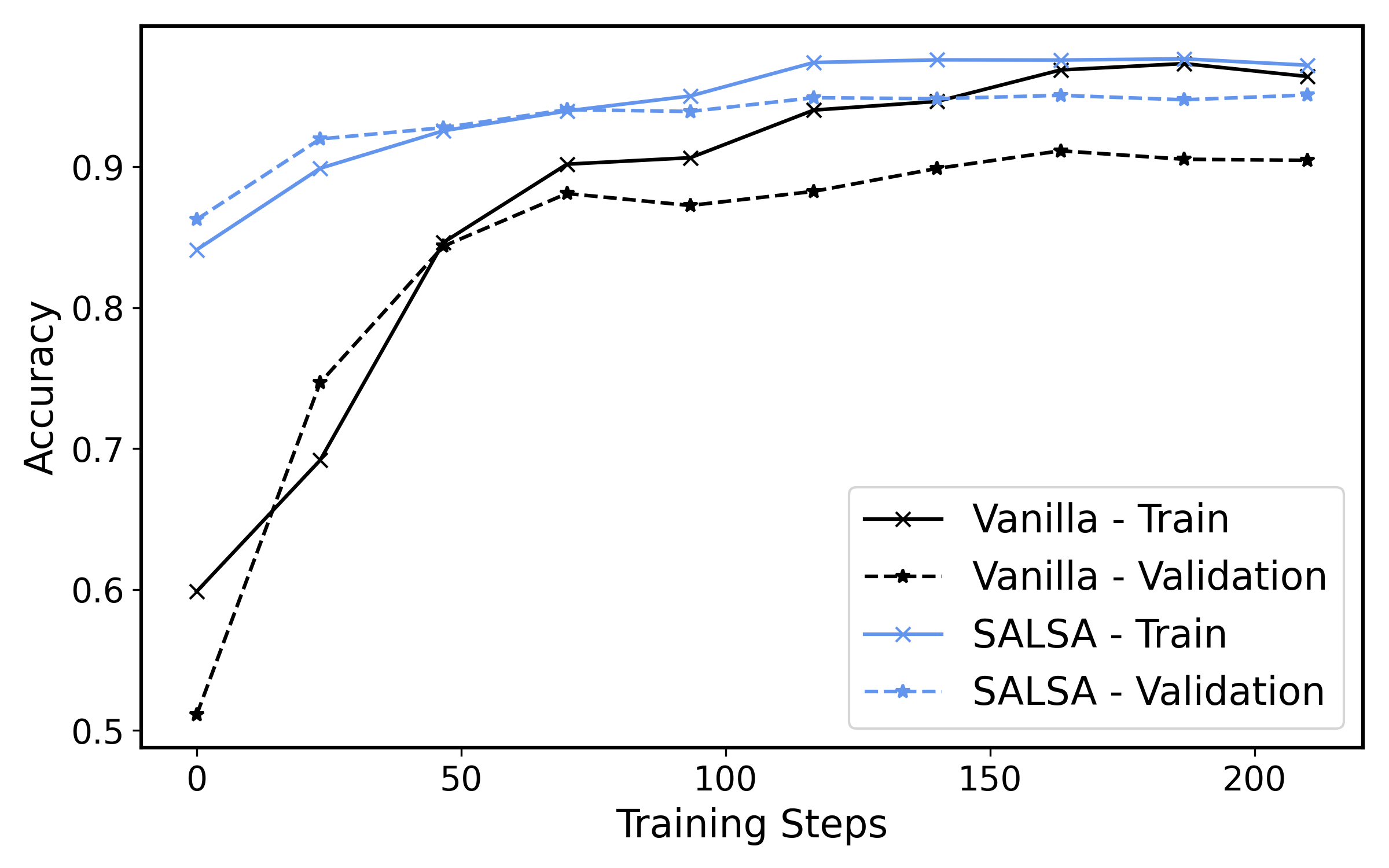}
    \caption{Convergence comparison between SALSA and Vanilla fine-tuning on RTE \citep{rte}. SALSA achieves faster convergence with higher accuracy on both training and validation sets, indicating better generalization and training efficiency.}
    \label{fig:convergence_1}
\end{figure}

\noindent\textbf{Efficient Optimization and Convergence.}\quad
To evaluate optimization efficiency, we compared SALSA to standard fine-tuning, where a linear classification head is added atop the base LLM’s final token output. Both methods were trained with identical hyperparameters to minimize cross-entropy loss. As shown in Figure~\ref{fig:convergence_1}, SALSA achieves faster convergence and consistently higher training and validation accuracy across steps. These results underscore its ability to reduce training time while improving generalization, making it well-suited for resource-constrained settings.

\section{Conclusion} 

SALSA exhibits consistent performance across its pipeline. As shown in Tables~\ref{tab:comparison_to_other_methods}, it achieves strong zero-shot results even without tuning. With fine-tuning, SALSA improves further without the instability often seen in cold-start training (Figure~\ref{fig:convergence_1}). It also reaches state-of-the-art accuracy across diverse tasks—sentiment analysis, medical QA, and hate speech detection—demonstrating broad applicability and strong generalization (Tables~\ref{tab:other_benchmarks}).

By reducing classification to a single forward pass, SALSA enables high-throughput use of large models, offering a more efficient alternative to generation-based approaches. Its use of LoRA adapters also preserves the base model’s capabilities for other LLM tasks.

While prompt design remains partly empirical, our ablation study (Section~\ref{sensetivity}) shows that fine-tuning mitigates label-token sensitivity. SALSA further supports regression tasks via discrete class ensembles (Section~\ref{discrete_to_cont}), extending its scope.

Future directions include systematic prompt optimization, adaptive thresholding, and unified extensions for multi-label and multi-task settings (Section~\ref{future}). Overall, SALSA offers a flexible and efficient framework for robust, general-purpose classification with generative LLMs.

\section{Limitations}
One key limitation of SALSA is its reliance on accessing the internal logit distribution of large language models (LLMs), which restricts its use to models or third-party services that expose such information. Additionally, the structured prompt design used to map classes to single tokens may not be applicable in all scenarios, particularly in tasks with more complex or nuanced label representations. Another concern is model contamination. Since we have no control over the data used to train the underlying LLM there is the possibility that some test examples may have been inadvertently incorporated during unsupervised training. Finally, SALSA inherits the biases and ethical concerns of its underlying LLM. As these models are trained on large-scale web corpora, they may encode and propagate societal biases, necessitating responsible use in real-world applications.

\bibliographystyle{acl_natbib}
\bibliography{anthology,custom}

\begin{thebibliography}{42}
\providecommand{\natexlab}[1]{#1}

\bibitem[{Ben~Zaken et~al.(2022)Ben~Zaken, Goldberg, and Ravfogel}]{ben-zaken-etal-2022-bitfit}
Elad Ben~Zaken, Yoav Goldberg, and Shauli Ravfogel. 2022.
\newblock \href {https://doi.org/10.18653/v1/2022.acl-short.1} {{B}it{F}it: Simple parameter-efficient fine-tuning for transformer-based masked language-models}.
\newblock In \emph{Proceedings of the 60th Annual Meeting of the Association for Computational Linguistics (Volume 2: Short Papers)}, pages 1--9, Dublin, Ireland. Association for Computational Linguistics.

\bibitem[{Bowman et~al.(2015)Bowman, Angeli, Potts, and Manning}]{mnli}
Samuel~R. Bowman, Gabor Angeli, Christopher Potts, and Christopher~D. Manning. 2015.
\newblock \href {https://doi.org/10.18653/v1/D15-1075} {A large annotated corpus for learning natural language inference}.
\newblock In \emph{Proceedings of the 2015 Conference on Empirical Methods in Natural Language Processing}, pages 632--642, Lisbon, Portugal. Association for Computational Linguistics.

\bibitem[{Brown et~al.(2020)Brown, Mann, Ryder, Subbiah, Kaplan, Dhariwal, Neelakantan, Shyam, Sastry, Askell, Agarwal, Herbert-Voss, Krueger, Henighan, Child, Ramesh, Ziegler, Wu, Winter, Hesse, Chen, Sigler, Litwin, Gray, Chess, Clark, Berner, McCandlish, Radford, Sutskever, and Amodei}]{brown2020language}
Tom Brown, Benjamin Mann, Nick Ryder, Melanie Subbiah, Jared~D Kaplan, Prafulla Dhariwal, Arvind Neelakantan, Pranav Shyam, Girish Sastry, Amanda Askell, Sandhini Agarwal, Ariel Herbert-Voss, Gretchen Krueger, Tom Henighan, Rewon Child, Aditya Ramesh, Daniel Ziegler, Jeffrey Wu, Clemens Winter, Chris Hesse, Mark Chen, Eric Sigler, Mateusz Litwin, Scott Gray, Benjamin Chess, Jack Clark, Christopher Berner, Sam McCandlish, Alec Radford, Ilya Sutskever, and Dario Amodei. 2020.
\newblock \href {https://proceedings.neurips.cc/paper_files/paper/2020/file/1457c0d6bfcb4967418bfb8ac142f64a-Paper.pdf} {Language models are few-shot learners}.
\newblock In \emph{Advances in Neural Information Processing Systems}, volume~33, pages 1877--1901. Curran Associates, Inc.

\bibitem[{Bucher and Martini(2024)}]{bucher2024fine}
Martin Juan~Jos{\'e} Bucher and Marco Martini. 2024.
\newblock Fine-tuned 'small' llms (still) significantly outperform zero-shot generative ai models in text classification.
\newblock \emph{arXiv preprint arXiv:2406.08660}.

\bibitem[{Cao et~al.(2021)Cao, Lin, Han, Sun, Yan, Liao, Xue, and Xu}]{cao-etal-2021-knowledgeable}
Boxi Cao, Hongyu Lin, Xianpei Han, Le~Sun, Lingyong Yan, Meng Liao, Tong Xue, and Jin Xu. 2021.
\newblock \href {https://doi.org/10.18653/v1/2021.acl-long.146} {Knowledgeable or educated guess? revisiting language models as knowledge bases}.
\newblock In \emph{Proceedings of the 59th Annual Meeting of the Association for Computational Linguistics and the 11th International Joint Conference on Natural Language Processing (Volume 1: Long Papers)}, pages 1860--1874, Online. Association for Computational Linguistics.

\bibitem[{Cer et~al.(2017)Cer, Diab, Agirre, Lopez-Gazpio, and Specia}]{cer-etal-2017-semeval}
Daniel Cer, Mona Diab, Eneko Agirre, I{\~n}igo Lopez-Gazpio, and Lucia Specia. 2017.
\newblock \href {https://doi.org/10.18653/v1/S17-2001} {{S}em{E}val-2017 task 1: Semantic textual similarity multilingual and crosslingual focused evaluation}.
\newblock In \emph{Proceedings of the 11th International Workshop on Semantic Evaluation ({S}em{E}val-2017)}, pages 1--14, Vancouver, Canada. Association for Computational Linguistics.

\bibitem[{Dagan et~al.(2005)Dagan, Glickman, and Magnini}]{rte}
Ido Dagan, Oren Glickman, and Bernardo Magnini. 2005.
\newblock The pascal recognising textual entailment challenge.
\newblock In \emph{Machine Learning Challenges Workshop}, pages 177--190. Springer.

\bibitem[{Devlin et~al.(2019)Devlin, Chang, Lee, and Toutanova}]{devlin-etal-2019-bert}
Jacob Devlin, Ming-Wei Chang, Kenton Lee, and Kristina Toutanova. 2019.
\newblock \href {https://doi.org/10.18653/v1/N19-1423} {{BERT}: Pre-training of deep bidirectional transformers for language understanding}.
\newblock In \emph{Proceedings of the 2019 Conference of the North {A}merican Chapter of the Association for Computational Linguistics: Human Language Technologies, Volume 1 (Long and Short Papers)}, pages 4171--4186, Minneapolis, Minnesota. Association for Computational Linguistics.

\bibitem[{Dolan and Brockett(2005)}]{mrpc}
William~B Dolan and Chris Brockett. 2005.
\newblock Automatically constructing a corpus of sentential paraphrases.
\newblock In \emph{Proceedings of the Third International Workshop on Paraphrasing (IWP2005)}.

\bibitem[{Grattafiori et~al.(2024)Grattafiori, Dubey, Jauhri, Pandey, Kadian, Al-Dahle, Letman, Mathur, Schelten, Vaughan, Yang, Fan, Goyal, Hartshorn, Yang, Mitra, Sravankumar, Korenev, Hinsvark, Rao, Zhang, Rodriguez, Gregerson, Spataru, Roziere, Biron, Tang, Chern, Caucheteux, Nayak, Bi, Marra, McConnell, Keller, Touret, Wu, Wong, Ferrer, Nikolaidis, Allonsius, Song, Pintz, Livshits, Wyatt, Esiobu, Choudhary, Mahajan, Garcia-Olano, Perino, Hupkes, Lakomkin, AlBadawy, Lobanova, Dinan, Smith, Radenovic, Guzmán, Zhang, Synnaeve, Lee, Anderson, Thattai, Nail, Mialon, Pang, Cucurell, Nguyen, Korevaar, Xu, Touvron, Zarov, Ibarra, Kloumann, Misra, Evtimov, Zhang, Copet, Lee, Geffert, Vranes, Park, Mahadeokar, Shah, van~der Linde, Billock, Hong, Lee, Fu, Chi, Huang, Liu, Wang, Yu, Bitton, Spisak, Park, Rocca, Johnstun, Saxe, Jia, Alwala, Prasad, Upasani, Plawiak, Li, Heafield, Stone, El-Arini, Iyer, Malik, Chiu, Bhalla, Lakhotia, Rantala-Yeary, van~der Maaten, Chen, Tan, Jenkins, Martin, Madaan, Malo, Blecher,
  Landzaat, de~Oliveira, Muzzi, Pasupuleti, Singh, Paluri, Kardas, Tsimpoukelli, Oldham, Rita, Pavlova, Kambadur, Lewis, Si, Singh, Hassan, Goyal, Torabi, Bashlykov, Bogoychev, Chatterji, Zhang, Duchenne, Çelebi, Alrassy, Zhang, Li, Vasic, Weng, Bhargava, Dubal, Krishnan, Koura, Xu, He, Dong, Srinivasan, Ganapathy, Calderer, Cabral, Stojnic, Raileanu, Maheswari, Girdhar, Patel, Sauvestre, Polidoro, Sumbaly, Taylor, Silva, Hou, Wang, Hosseini, Chennabasappa, Singh, Bell, Kim, Edunov, Nie, Narang, Raparthy, Shen, Wan, Bhosale, Zhang, Vandenhende, Batra, Whitman, Sootla, Collot, Gururangan, Borodinsky, Herman, Fowler, Sheasha, Georgiou, Scialom, Speckbacher, Mihaylov, Xiao, Karn, Goswami, Gupta, Ramanathan, Kerkez, Gonguet, Do, Vogeti, Albiero, Petrovic, Chu, Xiong, Fu, Meers, Martinet, Wang, Wang, Tan, Xia, Xie, Jia, Wang, Goldschlag, Gaur, Babaei, Wen, Song, Zhang, Li, Mao, Coudert, Yan, Chen, Papakipos, Singh, Srivastava, Jain, Kelsey, Shajnfeld, Gangidi, Victoria, Goldstand, Menon, Sharma, Boesenberg,
  Baevski, Feinstein, Kallet, Sangani, Teo, Yunus, Lupu, Alvarado, Caples, Gu, Ho, Poulton, Ryan, Ramchandani, Dong, Franco, Goyal, Saraf, Chowdhury, Gabriel, Bharambe, Eisenman, Yazdan, James, Maurer, Leonhardi, Huang, Loyd, Paola, Paranjape, Liu, Wu, Ni, Hancock, Wasti, Spence, Stojkovic, Gamido, Montalvo, Parker, Burton, Mejia, Liu, Wang, Kim, Zhou, Hu, Chu, Cai, Tindal, Feichtenhofer, Gao, Civin, Beaty, Kreymer, Li, Adkins, Xu, Testuggine, David, Parikh, Liskovich, Foss, Wang, Le, Holland, Dowling, Jamil, Montgomery, Presani, Hahn, Wood, Le, Brinkman, Arcaute, Dunbar, Smothers, Sun, Kreuk, Tian, Kokkinos, Ozgenel, Caggioni, Kanayet, Seide, Florez, Schwarz, Badeer, Swee, Halpern, Herman, Sizov, Guangyi, Zhang, Lakshminarayanan, Inan, Shojanazeri, Zou, Wang, Zha, Habeeb, Rudolph, Suk, Aspegren, Goldman, Zhan, Damlaj, Molybog, Tufanov, Leontiadis, Veliche, Gat, Weissman, Geboski, Kohli, Lam, Asher, Gaya, Marcus, Tang, Chan, Zhen, Reizenstein, Teboul, Zhong, Jin, Yang, Cummings, Carvill, Shepard, McPhie,
  Torres, Ginsburg, Wang, Wu, U, Saxena, Khandelwal, Zand, Matosich, Veeraraghavan, Michelena, Li, Jagadeesh, Huang, Chawla, Huang, Chen, Garg, A, Silva, Bell, Zhang, Guo, Yu, Moshkovich, Wehrstedt, Khabsa, Avalani, Bhatt, Mankus, Hasson, Lennie, Reso, Groshev, Naumov, Lathi, Keneally, Liu, Seltzer, Valko, Restrepo, Patel, Vyatskov, Samvelyan, Clark, Macey, Wang, Hermoso, Metanat, Rastegari, Bansal, Santhanam, Parks, White, Bawa, Singhal, Egebo, Usunier, Mehta, Laptev, Dong, Cheng, Chernoguz, Hart, Salpekar, Kalinli, Kent, Parekh, Saab, Balaji, Rittner, Bontrager, Roux, Dollar, Zvyagina, Ratanchandani, Yuvraj, Liang, Alao, Rodriguez, Ayub, Murthy, Nayani, Mitra, Parthasarathy, Li, Hogan, Battey, Wang, Howes, Rinott, Mehta, Siby, Bondu, Datta, Chugh, Hunt, Dhillon, Sidorov, Pan, Mahajan, Verma, Yamamoto, Ramaswamy, Lindsay, Lindsay, Feng, Lin, Zha, Patil, Shankar, Zhang, Zhang, Wang, Agarwal, Sajuyigbe, Chintala, Max, Chen, Kehoe, Satterfield, Govindaprasad, Gupta, Deng, Cho, Virk, Subramanian, Choudhury,
  Goldman, Remez, Glaser, Best, Koehler, Robinson, Li, Zhang, Matthews, Chou, Shaked, Vontimitta, Ajayi, Montanez, Mohan, Kumar, Mangla, Ionescu, Poenaru, Mihailescu, Ivanov, Li, Wang, Jiang, Bouaziz, Constable, Tang, Wu, Wang, Wu, Gao, Kleinman, Chen, Hu, Jia, Qi, Li, Zhang, Zhang, Adi, Nam, Yu, Wang, Zhao, Hao, Qian, Li, He, Rait, DeVito, Rosnbrick, Wen, Yang, Zhao, and Ma}]{grattafiori2024llama3herdmodels}
Aaron Grattafiori, Abhimanyu Dubey, Abhinav Jauhri, Abhinav Pandey, Abhishek Kadian, Ahmad Al-Dahle, Aiesha Letman, Akhil Mathur, Alan Schelten, Alex Vaughan, Amy Yang, Angela Fan, Anirudh Goyal, Anthony Hartshorn, Aobo Yang, Archi Mitra, Archie Sravankumar, Artem Korenev, Arthur Hinsvark, Arun Rao, Aston Zhang, Aurelien Rodriguez, Austen Gregerson, Ava Spataru, Baptiste Roziere, Bethany Biron, Binh Tang, Bobbie Chern, Charlotte Caucheteux, Chaya Nayak, Chloe Bi, Chris Marra, Chris McConnell, Christian Keller, Christophe Touret, Chunyang Wu, Corinne Wong, Cristian~Canton Ferrer, Cyrus Nikolaidis, Damien Allonsius, Daniel Song, Danielle Pintz, Danny Livshits, Danny Wyatt, David Esiobu, Dhruv Choudhary, Dhruv Mahajan, Diego Garcia-Olano, Diego Perino, Dieuwke Hupkes, Egor Lakomkin, Ehab AlBadawy, Elina Lobanova, Emily Dinan, Eric~Michael Smith, Filip Radenovic, Francisco Guzmán, Frank Zhang, Gabriel Synnaeve, Gabrielle Lee, Georgia~Lewis Anderson, Govind Thattai, Graeme Nail, Gregoire Mialon, Guan Pang,
  Guillem Cucurell, Hailey Nguyen, Hannah Korevaar, Hu~Xu, Hugo Touvron, Iliyan Zarov, Imanol~Arrieta Ibarra, Isabel Kloumann, Ishan Misra, Ivan Evtimov, Jack Zhang, Jade Copet, Jaewon Lee, Jan Geffert, Jana Vranes, Jason Park, Jay Mahadeokar, Jeet Shah, Jelmer van~der Linde, Jennifer Billock, Jenny Hong, Jenya Lee, Jeremy Fu, Jianfeng Chi, Jianyu Huang, Jiawen Liu, Jie Wang, Jiecao Yu, Joanna Bitton, Joe Spisak, Jongsoo Park, Joseph Rocca, Joshua Johnstun, Joshua Saxe, Junteng Jia, Kalyan~Vasuden Alwala, Karthik Prasad, Kartikeya Upasani, Kate Plawiak, Ke~Li, Kenneth Heafield, Kevin Stone, Khalid El-Arini, Krithika Iyer, Kshitiz Malik, Kuenley Chiu, Kunal Bhalla, Kushal Lakhotia, Lauren Rantala-Yeary, Laurens van~der Maaten, Lawrence Chen, Liang Tan, Liz Jenkins, Louis Martin, Lovish Madaan, Lubo Malo, Lukas Blecher, Lukas Landzaat, Luke de~Oliveira, Madeline Muzzi, Mahesh Pasupuleti, Mannat Singh, Manohar Paluri, Marcin Kardas, Maria Tsimpoukelli, Mathew Oldham, Mathieu Rita, Maya Pavlova, Melanie Kambadur,
  Mike Lewis, Min Si, Mitesh~Kumar Singh, Mona Hassan, Naman Goyal, Narjes Torabi, Nikolay Bashlykov, Nikolay Bogoychev, Niladri Chatterji, Ning Zhang, Olivier Duchenne, Onur Çelebi, Patrick Alrassy, Pengchuan Zhang, Pengwei Li, Petar Vasic, Peter Weng, Prajjwal Bhargava, Pratik Dubal, Praveen Krishnan, Punit~Singh Koura, Puxin Xu, Qing He, Qingxiao Dong, Ragavan Srinivasan, Raj Ganapathy, Ramon Calderer, Ricardo~Silveira Cabral, Robert Stojnic, Roberta Raileanu, Rohan Maheswari, Rohit Girdhar, Rohit Patel, Romain Sauvestre, Ronnie Polidoro, Roshan Sumbaly, Ross Taylor, Ruan Silva, Rui Hou, Rui Wang, Saghar Hosseini, Sahana Chennabasappa, Sanjay Singh, Sean Bell, Seohyun~Sonia Kim, Sergey Edunov, Shaoliang Nie, Sharan Narang, Sharath Raparthy, Sheng Shen, Shengye Wan, Shruti Bhosale, Shun Zhang, Simon Vandenhende, Soumya Batra, Spencer Whitman, Sten Sootla, Stephane Collot, Suchin Gururangan, Sydney Borodinsky, Tamar Herman, Tara Fowler, Tarek Sheasha, Thomas Georgiou, Thomas Scialom, Tobias Speckbacher,
  Todor Mihaylov, Tong Xiao, Ujjwal Karn, Vedanuj Goswami, Vibhor Gupta, Vignesh Ramanathan, Viktor Kerkez, Vincent Gonguet, Virginie Do, Vish Vogeti, Vítor Albiero, Vladan Petrovic, Weiwei Chu, Wenhan Xiong, Wenyin Fu, Whitney Meers, Xavier Martinet, Xiaodong Wang, Xiaofang Wang, Xiaoqing~Ellen Tan, Xide Xia, Xinfeng Xie, Xuchao Jia, Xuewei Wang, Yaelle Goldschlag, Yashesh Gaur, Yasmine Babaei, Yi~Wen, Yiwen Song, Yuchen Zhang, Yue Li, Yuning Mao, Zacharie~Delpierre Coudert, Zheng Yan, Zhengxing Chen, Zoe Papakipos, Aaditya Singh, Aayushi Srivastava, Abha Jain, Adam Kelsey, Adam Shajnfeld, Adithya Gangidi, Adolfo Victoria, Ahuva Goldstand, Ajay Menon, Ajay Sharma, Alex Boesenberg, Alexei Baevski, Allie Feinstein, Amanda Kallet, Amit Sangani, Amos Teo, Anam Yunus, Andrei Lupu, Andres Alvarado, Andrew Caples, Andrew Gu, Andrew Ho, Andrew Poulton, Andrew Ryan, Ankit Ramchandani, Annie Dong, Annie Franco, Anuj Goyal, Aparajita Saraf, Arkabandhu Chowdhury, Ashley Gabriel, Ashwin Bharambe, Assaf Eisenman, Azadeh
  Yazdan, Beau James, Ben Maurer, Benjamin Leonhardi, Bernie Huang, Beth Loyd, Beto~De Paola, Bhargavi Paranjape, Bing Liu, Bo~Wu, Boyu Ni, Braden Hancock, Bram Wasti, Brandon Spence, Brani Stojkovic, Brian Gamido, Britt Montalvo, Carl Parker, Carly Burton, Catalina Mejia, Ce~Liu, Changhan Wang, Changkyu Kim, Chao Zhou, Chester Hu, Ching-Hsiang Chu, Chris Cai, Chris Tindal, Christoph Feichtenhofer, Cynthia Gao, Damon Civin, Dana Beaty, Daniel Kreymer, Daniel Li, David Adkins, David Xu, Davide Testuggine, Delia David, Devi Parikh, Diana Liskovich, Didem Foss, Dingkang Wang, Duc Le, Dustin Holland, Edward Dowling, Eissa Jamil, Elaine Montgomery, Eleonora Presani, Emily Hahn, Emily Wood, Eric-Tuan Le, Erik Brinkman, Esteban Arcaute, Evan Dunbar, Evan Smothers, Fei Sun, Felix Kreuk, Feng Tian, Filippos Kokkinos, Firat Ozgenel, Francesco Caggioni, Frank Kanayet, Frank Seide, Gabriela~Medina Florez, Gabriella Schwarz, Gada Badeer, Georgia Swee, Gil Halpern, Grant Herman, Grigory Sizov, Guangyi, Zhang, Guna
  Lakshminarayanan, Hakan Inan, Hamid Shojanazeri, Han Zou, Hannah Wang, Hanwen Zha, Haroun Habeeb, Harrison Rudolph, Helen Suk, Henry Aspegren, Hunter Goldman, Hongyuan Zhan, Ibrahim Damlaj, Igor Molybog, Igor Tufanov, Ilias Leontiadis, Irina-Elena Veliche, Itai Gat, Jake Weissman, James Geboski, James Kohli, Janice Lam, Japhet Asher, Jean-Baptiste Gaya, Jeff Marcus, Jeff Tang, Jennifer Chan, Jenny Zhen, Jeremy Reizenstein, Jeremy Teboul, Jessica Zhong, Jian Jin, Jingyi Yang, Joe Cummings, Jon Carvill, Jon Shepard, Jonathan McPhie, Jonathan Torres, Josh Ginsburg, Junjie Wang, Kai Wu, Kam~Hou U, Karan Saxena, Kartikay Khandelwal, Katayoun Zand, Kathy Matosich, Kaushik Veeraraghavan, Kelly Michelena, Keqian Li, Kiran Jagadeesh, Kun Huang, Kunal Chawla, Kyle Huang, Lailin Chen, Lakshya Garg, Lavender A, Leandro Silva, Lee Bell, Lei Zhang, Liangpeng Guo, Licheng Yu, Liron Moshkovich, Luca Wehrstedt, Madian Khabsa, Manav Avalani, Manish Bhatt, Martynas Mankus, Matan Hasson, Matthew Lennie, Matthias Reso, Maxim
  Groshev, Maxim Naumov, Maya Lathi, Meghan Keneally, Miao Liu, Michael~L. Seltzer, Michal Valko, Michelle Restrepo, Mihir Patel, Mik Vyatskov, Mikayel Samvelyan, Mike Clark, Mike Macey, Mike Wang, Miquel~Jubert Hermoso, Mo~Metanat, Mohammad Rastegari, Munish Bansal, Nandhini Santhanam, Natascha Parks, Natasha White, Navyata Bawa, Nayan Singhal, Nick Egebo, Nicolas Usunier, Nikhil Mehta, Nikolay~Pavlovich Laptev, Ning Dong, Norman Cheng, Oleg Chernoguz, Olivia Hart, Omkar Salpekar, Ozlem Kalinli, Parkin Kent, Parth Parekh, Paul Saab, Pavan Balaji, Pedro Rittner, Philip Bontrager, Pierre Roux, Piotr Dollar, Polina Zvyagina, Prashant Ratanchandani, Pritish Yuvraj, Qian Liang, Rachad Alao, Rachel Rodriguez, Rafi Ayub, Raghotham Murthy, Raghu Nayani, Rahul Mitra, Rangaprabhu Parthasarathy, Raymond Li, Rebekkah Hogan, Robin Battey, Rocky Wang, Russ Howes, Ruty Rinott, Sachin Mehta, Sachin Siby, Sai~Jayesh Bondu, Samyak Datta, Sara Chugh, Sara Hunt, Sargun Dhillon, Sasha Sidorov, Satadru Pan, Saurabh Mahajan,
  Saurabh Verma, Seiji Yamamoto, Sharadh Ramaswamy, Shaun Lindsay, Shaun Lindsay, Sheng Feng, Shenghao Lin, Shengxin~Cindy Zha, Shishir Patil, Shiva Shankar, Shuqiang Zhang, Shuqiang Zhang, Sinong Wang, Sneha Agarwal, Soji Sajuyigbe, Soumith Chintala, Stephanie Max, Stephen Chen, Steve Kehoe, Steve Satterfield, Sudarshan Govindaprasad, Sumit Gupta, Summer Deng, Sungmin Cho, Sunny Virk, Suraj Subramanian, Sy~Choudhury, Sydney Goldman, Tal Remez, Tamar Glaser, Tamara Best, Thilo Koehler, Thomas Robinson, Tianhe Li, Tianjun Zhang, Tim Matthews, Timothy Chou, Tzook Shaked, Varun Vontimitta, Victoria Ajayi, Victoria Montanez, Vijai Mohan, Vinay~Satish Kumar, Vishal Mangla, Vlad Ionescu, Vlad Poenaru, Vlad~Tiberiu Mihailescu, Vladimir Ivanov, Wei Li, Wenchen Wang, Wenwen Jiang, Wes Bouaziz, Will Constable, Xiaocheng Tang, Xiaojian Wu, Xiaolan Wang, Xilun Wu, Xinbo Gao, Yaniv Kleinman, Yanjun Chen, Ye~Hu, Ye~Jia, Ye~Qi, Yenda Li, Yilin Zhang, Ying Zhang, Yossi Adi, Youngjin Nam, Yu, Wang, Yu~Zhao, Yuchen Hao, Yundi
  Qian, Yunlu Li, Yuzi He, Zach Rait, Zachary DeVito, Zef Rosnbrick, Zhaoduo Wen, Zhenyu Yang, Zhiwei Zhao, and Zhiyu Ma. 2024.
\newblock \href {https://arxiv.org/abs/2407.21783} {The llama 3 herd of models}.
\newblock \emph{Preprint}, arXiv:2407.21783.

\bibitem[{Hu et~al.(2021)Hu, Shen, Wallis, Allen-Zhu, Li, Wang, Wang, and Chen}]{hu2021lora}
Edward~J Hu, Yelong Shen, Phillip Wallis, Zeyuan Allen-Zhu, Yuanzhi Li, Shean Wang, Lu~Wang, and Weizhu Chen. 2021.
\newblock Lora: Low-rank adaptation of large language models.
\newblock \emph{arXiv preprint arXiv:2106.09685}.

\bibitem[{Iyer et~al.(2017)Iyer, Dandekar, and Csernai}]{qqp}
Shankar Iyer, Nikhil Dandekar, and Kornel Csernai. 2017.
\newblock \href {https://data.quora.com/First-Quora-Dataset-Release-Question-Pairs} {First quora dataset release: Question pairs}.

\bibitem[{Kim et~al.(2022)Kim, Lee, and Sohn}]{kim2022hatespeechmaskedrationale}
Jiyun Kim, Byounghan Lee, and Kyung-Ah Sohn. 2022.
\newblock \href {https://arxiv.org/abs/2211.00243} {Why is it hate speech? masked rationale prediction for explainable hate speech detection}.
\newblock \emph{Preprint}, arXiv:2211.00243.

\bibitem[{Kim(2014)}]{kim-2014-convolutional}
Yoon Kim. 2014.
\newblock \href {https://doi.org/10.3115/v1/D14-1181} {Convolutional neural networks for sentence classification}.
\newblock In \emph{Proceedings of the 2014 Conference on Empirical Methods in Natural Language Processing ({EMNLP})}, pages 1746--1751, Doha, Qatar. Association for Computational Linguistics.

\bibitem[{Kingma(2014)}]{kingma2014adam}
Diederik~P Kingma. 2014.
\newblock Adam: A method for stochastic optimization.
\newblock \emph{arXiv preprint arXiv:1412.6980}.

\bibitem[{Lan et~al.(2019)Lan, Chen, Goodman, Gimpel, Sharma, and Soricut}]{DBLP:journals/corr/abs-1909-11942}
Zhenzhong Lan, Mingda Chen, Sebastian Goodman, Kevin Gimpel, Piyush Sharma, and Radu Soricut. 2019.
\newblock \href {https://arxiv.org/abs/1909.11942} {{ALBERT:} {A} lite {BERT} for self-supervised learning of language representations}.
\newblock \emph{CoRR}, abs/1909.11942.

\bibitem[{Liu et~al.(2019)Liu, Ott, Goyal, Du, Joshi, Chen, Levy, Lewis, Zettlemoyer, and Stoyanov}]{liu2019roberta}
Yinhan Liu, Myle Ott, Naman Goyal, Jingfei Du, Mandar Joshi, Danqi Chen, Omer Levy, Mike Lewis, Luke Zettlemoyer, and Veselin Stoyanov. 2019.
\newblock {RoBERTa}: A robustly optimized {BERT} pretraining approach.
\newblock \emph{arXiv preprint arXiv:1907.11692}.

\bibitem[{Maas et~al.(2011)Maas, Daly, Pham, Huang, Ng, and Potts}]{maas-EtAl:2011:ACL-HLT2011}
Andrew~L. Maas, Raymond~E. Daly, Peter~T. Pham, Dan Huang, Andrew~Y. Ng, and Christopher Potts. 2011.
\newblock \href {http://www.aclweb.org/anthology/P11-1015} {Learning word vectors for sentiment analysis}.
\newblock In \emph{Proceedings of the 49th Annual Meeting of the Association for Computational Linguistics: Human Language Technologies}, pages 142--150, Portland, Oregon, USA. Association for Computational Linguistics.

\bibitem[{Mathew et~al.(2020)Mathew, Saha, Yimam, Biemann, Goyal, and Mukherjee}]{DBLP:journals/corr/abs-2012-10289}
Binny Mathew, Punyajoy Saha, Seid~Muhie Yimam, Chris Biemann, Pawan Goyal, and Animesh Mukherjee. 2020.
\newblock \href {https://arxiv.org/abs/2012.10289} {Hatexplain: {A} benchmark dataset for explainable hate speech detection}.
\newblock \emph{CoRR}, abs/2012.10289.

\bibitem[{Mikolov et~al.(2013)Mikolov, Chen, Corrado, and Dean}]{DBLP:journals/corr/abs-1301-3781}
Tom{\'{a}}s Mikolov, Kai Chen, Greg Corrado, and Jeffrey Dean. 2013.
\newblock \href {http://arxiv.org/abs/1301.3781} {Efficient estimation of word representations in vector space}.
\newblock In \emph{1st International Conference on Learning Representations, {ICLR} 2013, Scottsdale, Arizona, USA, May 2-4, 2013, Workshop Track Proceedings}.

\bibitem[{Nori et~al.(2023)Nori, King, McKinney, Carignan, and Horvitz}]{nori2023capabilitiesgpt4medicalchallenge}
Harsha Nori, Nicholas King, Scott~Mayer McKinney, Dean Carignan, and Eric Horvitz. 2023.
\newblock \href {https://arxiv.org/abs/2303.13375} {Capabilities of gpt-4 on medical challenge problems}.
\newblock \emph{Preprint}, arXiv:2303.13375.

\bibitem[{Pal et~al.(2022)Pal, Umapathi, and Sankarasubbu}]{pal2022medmcqalargescalemultisubject}
Ankit Pal, Logesh~Kumar Umapathi, and Malaikannan Sankarasubbu. 2022.
\newblock \href {https://arxiv.org/abs/2203.14371} {Medmcqa : A large-scale multi-subject multi-choice dataset for medical domain question answering}.
\newblock \emph{Preprint}, arXiv:2203.14371.

\bibitem[{Pawar et~al.(2024)Pawar, Ramrakhiyani, Sinha, Apte, and Palshikar}]{pawar-etal-2024-generate}
Sachin Pawar, Nitin Ramrakhiyani, Anubhav Sinha, Manoj Apte, and Girish Palshikar. 2024.
\newblock \href {https://aclanthology.org/2024.findings-eacl.74/} {Why generate when you can discriminate? a novel technique for text classification using language models}.
\newblock In \emph{Findings of the Association for Computational Linguistics: EACL 2024}, pages 1099--1114, St. Julian{'}s, Malta. Association for Computational Linguistics.

\bibitem[{Peters et~al.(2018)Peters, Neumann, Iyyer, Gardner, Clark, Lee, and Zettlemoyer}]{peters-etal-2018-deep}
Matthew~E. Peters, Mark Neumann, Mohit Iyyer, Matt Gardner, Christopher Clark, Kenton Lee, and Luke Zettlemoyer. 2018.
\newblock \href {https://doi.org/10.18653/v1/N18-1202} {Deep contextualized word representations}.
\newblock In \emph{Proceedings of the 2018 Conference of the North {A}merican Chapter of the Association for Computational Linguistics: Human Language Technologies, Volume 1 (Long Papers)}, pages 2227--2237, New Orleans, Louisiana. Association for Computational Linguistics.

\bibitem[{Petroni et~al.(2019)Petroni, Rockt{\"a}schel, Lewis, Bakhtin, Wu, Miller, and Riedel}]{petroni2019language}
Fabio Petroni, Tim Rockt{\"a}schel, Patrick Lewis, Anton Bakhtin, Yuxiang Wu, Alexander~H Miller, and Sebastian Riedel. 2019.
\newblock Language models as knowledge bases?
\newblock In \emph{Proceedings of the 2019 Conference on Empirical Methods in Natural Language Processing and the 9th International Joint Conference on Natural Language Processing (EMNLP-IJCNLP)}, pages 2463--2473. Association for Computational Linguistics.

\bibitem[{Raffel et~al.(2020)Raffel, Shazeer, Roberts, Lee, Narang, Matena, Zhou, Li, and Liu}]{t5}
Colin Raffel, Noam Shazeer, Adam Roberts, Katherine Lee, Sharan Narang, Michael Matena, Yanqi Zhou, Wei Li, and Peter~J. Liu. 2020.
\newblock Exploring the limits of transfer learning with a unified text-to-text transformer.
\newblock \emph{Journal of Machine Learning Research}, 21(1):1--67.

\bibitem[{Rajpurkar et~al.(2016)Rajpurkar, Zhang, Lopyrev, and Liang}]{qnli}
Pranav Rajpurkar, Jian Zhang, Konstantin Lopyrev, and Percy Liang. 2016.
\newblock \href {https://doi.org/10.18653/v1/D16-1264} {{SQ}u{AD}: 100,000+ questions for machine comprehension of text}.
\newblock In \emph{Proceedings of the 2016 Conference on Empirical Methods in Natural Language Processing}, pages 2383--2392, Austin, Texas. Association for Computational Linguistics.

\bibitem[{Robinson et~al.(2022)Robinson, Rytting, and Wingate}]{robinson2022leveraging}
Joshua Robinson, Christopher~Michael Rytting, and David Wingate. 2022.
\newblock Leveraging large language models for multiple choice question answering.
\newblock \emph{arXiv preprint arXiv:2210.12353}.

\bibitem[{Romanov and Shivade(2018)}]{romanov-shivade-2018-lessons}
Alexey Romanov and Chaitanya Shivade. 2018.
\newblock \href {https://doi.org/10.18653/v1/D18-1187} {Lessons from natural language inference in the clinical domain}.
\newblock In \emph{Proceedings of the 2018 Conference on Empirical Methods in Natural Language Processing}, pages 1586--1596, Brussels, Belgium. Association for Computational Linguistics.

\bibitem[{Socher et~al.(2013)Socher, Perelygin, Wu, Chuang, Manning, Ng, and Potts}]{sst2}
Richard Socher, Alex Perelygin, Jean Wu, Jason Chuang, Christopher~D Manning, Andrew Ng, and Christopher Potts. 2013.
\newblock Recursive deep models for semantic compositionality over a sentiment treebank.
\newblock In \emph{Proceedings of the 2013 conference on empirical methods in natural language processing}, pages 1631--1642.

\bibitem[{Team et~al.(2024)Team, Mesnard, Hardin, Dadashi, Bhupatiraju, Pathak, Sifre, Rivière, Kale, Love, Tafti, Hussenot, Sessa, Chowdhery, Roberts, Barua, Botev, Castro-Ros, Slone, Héliou, Tacchetti, Bulanova, Paterson, Tsai, Shahriari, Lan, Choquette-Choo, Crepy, Cer, Ippolito, Reid, Buchatskaya, Ni, Noland, Yan, Tucker, Muraru, Rozhdestvenskiy, Michalewski, Tenney, Grishchenko, Austin, Keeling, Labanowski, Lespiau, Stanway, Brennan, Chen, Ferret, Chiu, Mao-Jones, Lee, Yu, Millican, Sjoesund, Lee, Dixon, Reid, Mikuła, Wirth, Sharman, Chinaev, Thain, Bachem, Chang, Wahltinez, Bailey, Michel, Yotov, Chaabouni, Comanescu, Jana, Anil, McIlroy, Liu, Mullins, Smith, Borgeaud, Girgin, Douglas, Pandya, Shakeri, De, Klimenko, Hennigan, Feinberg, Stokowiec, hui Chen, Ahmed, Gong, Warkentin, Peran, Giang, Farabet, Vinyals, Dean, Kavukcuoglu, Hassabis, Ghahramani, Eck, Barral, Pereira, Collins, Joulin, Fiedel, Senter, Andreev, and Kenealy}]{gemmateam2024gemmaopenmodelsbased}
Gemma Team, Thomas Mesnard, Cassidy Hardin, Robert Dadashi, Surya Bhupatiraju, Shreya Pathak, Laurent Sifre, Morgane Rivière, Mihir~Sanjay Kale, Juliette Love, Pouya Tafti, Léonard Hussenot, Pier~Giuseppe Sessa, Aakanksha Chowdhery, Adam Roberts, Aditya Barua, Alex Botev, Alex Castro-Ros, Ambrose Slone, Amélie Héliou, Andrea Tacchetti, Anna Bulanova, Antonia Paterson, Beth Tsai, Bobak Shahriari, Charline~Le Lan, Christopher~A. Choquette-Choo, Clément Crepy, Daniel Cer, Daphne Ippolito, David Reid, Elena Buchatskaya, Eric Ni, Eric Noland, Geng Yan, George Tucker, George-Christian Muraru, Grigory Rozhdestvenskiy, Henryk Michalewski, Ian Tenney, Ivan Grishchenko, Jacob Austin, James Keeling, Jane Labanowski, Jean-Baptiste Lespiau, Jeff Stanway, Jenny Brennan, Jeremy Chen, Johan Ferret, Justin Chiu, Justin Mao-Jones, Katherine Lee, Kathy Yu, Katie Millican, Lars~Lowe Sjoesund, Lisa Lee, Lucas Dixon, Machel Reid, Maciej Mikuła, Mateo Wirth, Michael Sharman, Nikolai Chinaev, Nithum Thain, Olivier Bachem,
  Oscar Chang, Oscar Wahltinez, Paige Bailey, Paul Michel, Petko Yotov, Rahma Chaabouni, Ramona Comanescu, Reena Jana, Rohan Anil, Ross McIlroy, Ruibo Liu, Ryan Mullins, Samuel~L Smith, Sebastian Borgeaud, Sertan Girgin, Sholto Douglas, Shree Pandya, Siamak Shakeri, Soham De, Ted Klimenko, Tom Hennigan, Vlad Feinberg, Wojciech Stokowiec, Yu~hui Chen, Zafarali Ahmed, Zhitao Gong, Tris Warkentin, Ludovic Peran, Minh Giang, Clément Farabet, Oriol Vinyals, Jeff Dean, Koray Kavukcuoglu, Demis Hassabis, Zoubin Ghahramani, Douglas Eck, Joelle Barral, Fernando Pereira, Eli Collins, Armand Joulin, Noah Fiedel, Evan Senter, Alek Andreev, and Kathleen Kenealy. 2024.
\newblock \href {https://arxiv.org/abs/2403.08295} {Gemma: Open models based on gemini research and technology}.
\newblock \emph{Preprint}, arXiv:2403.08295.

\bibitem[{Team(2022)}]{TULRv6}
Microsoft~Turing Team. 2022.
\newblock \href {https://blogs.bing.com/search-quality-insights/october-2022/Microsoft-Turing-Universal-Language-Representation-model%2C-T-ULRv6%2C-tops-both-XTREME-and-GLUE-leaderb} {Microsoft turing universal language representation model (t-ulrv6)}.

\bibitem[{Tiwary and Zhou(2021)}]{TULRv5}
Saurabh Tiwary and Lidong Zhou. 2021.
\newblock \href {https://www.microsoft.com/en-us/research/blog/microsoft-turing-universal-language-representation-model-t-ulrv5-tops-xtreme-leaderboard-and-trains-100x-faster/} {Microsoft turing universal language representation model, t-ulrv5, tops xtreme leaderboard and trains 100x faster}.
\newblock \emph{Microsoft Research Blog}.

\bibitem[{Touvron et~al.(2023)Touvron, Lavril, Izacard, Martinet, Lachaux, Lacroix, Rozière, Goyal, Hambro, Azhar, Rodriguez, Joulin, Grave, and Lample}]{touvron2023llamaopenefficientfoundation}
Hugo Touvron, Thibaut Lavril, Gautier Izacard, Xavier Martinet, Marie-Anne Lachaux, Timothée Lacroix, Baptiste Rozière, Naman Goyal, Eric Hambro, Faisal Azhar, Aurelien Rodriguez, Armand Joulin, Edouard Grave, and Guillaume Lample. 2023.
\newblock \href {https://arxiv.org/abs/2302.13971} {Llama: Open and efficient foundation language models}.
\newblock \emph{Preprint}, arXiv:2302.13971.

\bibitem[{Vaswani et~al.(2017)Vaswani, Shazeer, Parmar, Uszkoreit, Jones, Gomez, Kaiser, and Polosukhin}]{NIPS2017_3f5ee243}
Ashish Vaswani, Noam Shazeer, Niki Parmar, Jakob Uszkoreit, Llion Jones, Aidan~N Gomez, \L~ukasz Kaiser, and Illia Polosukhin. 2017.
\newblock \href {https://proceedings.neurips.cc/paper_files/paper/2017/file/3f5ee243547dee91fbd053c1c4a845aa-Paper.pdf} {Attention is all you need}.
\newblock In \emph{Advances in Neural Information Processing Systems}, volume~30. Curran Associates, Inc.

\bibitem[{Wang et~al.(2018)Wang, Singh, Michael, Hill, Levy, and Bowman}]{glue}
Alex Wang, Amanpreet Singh, Julian Michael, Felix Hill, Omer Levy, and Samuel Bowman. 2018.
\newblock \href {https://doi.org/10.18653/v1/W18-5446} {{GLUE}: A multi-task benchmark and analysis platform for natural language understanding}.
\newblock In \emph{Proceedings of the 2018 {EMNLP} Workshop {B}lackbox{NLP}: Analyzing and Interpreting Neural Networks for {NLP}}, pages 353--355, Brussels, Belgium. Association for Computational Linguistics.

\bibitem[{Wei et~al.(2021)Wei, Bosma, Zhao, Guu, Yu, Lester, Du, Dai, and Le}]{wei2021finetuned}
Jason Wei, Maarten Bosma, Vincent~Y Zhao, Kelvin Guu, Adams~Wei Yu, Brian Lester, Nan Du, Andrew~M Dai, and Quoc~V Le. 2021.
\newblock Finetuned language models are zero-shot learners.
\newblock \emph{arXiv preprint arXiv:2109.01652}.

\bibitem[{Wei et~al.(2022)Wei, Wang, Schuurmans, Bosma, Xia, Chi, Le, Zhou et~al.}]{wei2022chain}
Jason Wei, Xuezhi Wang, Dale Schuurmans, Maarten Bosma, Fei Xia, Ed~Chi, Quoc~V Le, Denny Zhou, et~al. 2022.
\newblock Chain-of-thought prompting elicits reasoning in large language models.
\newblock \emph{Advances in neural information processing systems}, 35:24824--24837.

\bibitem[{Yang et~al.(2022)Yang, Chen, PourNejatian, Shin, Smith, Parisien, Compas, Martin, Flores, Zhang, Magoc, Harle, Lipori, Mitchell, Hogan, Shenkman, Bian, and Wu}]{yang2022gatortronlargeclinicallanguage}
Xi~Yang, Aokun Chen, Nima PourNejatian, Hoo~Chang Shin, Kaleb~E Smith, Christopher Parisien, Colin Compas, Cheryl Martin, Mona~G Flores, Ying Zhang, Tanja Magoc, Christopher~A Harle, Gloria Lipori, Duane~A Mitchell, William~R Hogan, Elizabeth~A Shenkman, Jiang Bian, and Yonghui Wu. 2022.
\newblock \href {https://arxiv.org/abs/2203.03540} {Gatortron: A large clinical language model to unlock patient information from unstructured electronic health records}.
\newblock \emph{Preprint}, arXiv:2203.03540.

\bibitem[{Yang et~al.(2019)Yang, Dai, Yang, Carbonell, Salakhutdinov, and Le}]{yang2019xlnet}
Zhilin Yang, Zihang Dai, Yiming Yang, Jaime Carbonell, Ruslan Salakhutdinov, and Quoc~V. Le. 2019.
\newblock Xlnet: Generalized autoregressive pretraining for language understanding.
\newblock In \emph{Proceedings of the 33rd International Conference on Neural Information Processing Systems (NeurIPS)}, Red Hook, NY, USA. Curran Associates Inc.

\bibitem[{Zhang et~al.(2015)Zhang, Zhao, and LeCun}]{yelp}
Xiang Zhang, Junbo Zhao, and Yann LeCun. 2015.
\newblock \href {https://proceedings.neurips.cc/paper_files/paper/2015/file/250cf8b51c773f3f8dc8b4be867a9a02-Paper.pdf} {Character-level convolutional networks for text classification}.
\newblock In \emph{Advances in Neural Information Processing Systems}, volume~28. Curran Associates, Inc.

\bibitem[{Zhong et~al.(2023)Zhong, Ding, Peng, Liu, Du, Shen, Zhan, and Tao}]{vega_v1}
Qihuang Zhong, Liang Ding, Keqin Peng, Juhua Liu, Bo~Du, Li~Shen, Yibing Zhan, and Dacheng Tao. 2023.
\newblock \href {https://arxiv.org/abs/2302.09268} {Bag of tricks for effective language model pretraining and downstream adaptation: A case study on glue}.
\newblock \emph{Preprint}, arXiv:2302.09268.

\end{thebibliography}

\clearpage
\appendix
\section{Appendices}
\label{sec:appendix}

\subsection{Datasets}\label{datasets_section}
We used multiple datasets to evaluate SALSA, focusing on text classification tasks.\\

\noindent\textbf{GLUE Benchmark.}\quad
We evaluated SALSA on a subset of tasks from the GLUE benchmark \cite{glue} and report both the task details and evaluation metrics. Specifically, we tested on the following tasks: the Stanford Sentiment Treebank (SST-2; \citet{sst2}), the Microsoft Research Paraphrase Corpus (MRPC; \citet{mrpc}), the Quora Question Pairs (QQP; \citet{qqp}), the Multi-Genre Natural Language Inference Corpus (MNLI; \citet{mnli}), the Stanford Question Answering Dataset (QNLI; \citet{qnli}), and Recognizing Textual Entailment (RTE; \citet{rte}).\\
 
\noindent\textbf{AG's News.}\quad
The AG's News dataset \citep{yelp} includes 120,000+ news articles across four categories (World, Sports, Business, Science/Technology), testing LLM robustness with diverse topics and journalistic tones.\\

\noindent\textbf{IMDb.}\quad
The IMDb data set \citep{maas-EtAl:2011:ACL-HLT2011} is a benchmark for binary sentiment analysis with positive or negative movie reviews, testing classification models on diverse styles of writing, topics, and sentiment intensities.\\

\noindent\textbf{Yelp-5.}\quad
The Yelp-5 dataset \citep{yelp}, used for multi-class sentiment analysis, contains customer reviews rated 1-5 stars, challenging models with varied review lengths, tones, and topics.\\

\noindent\textbf{HateXplain.}\quad
HateXplain \citep{DBLP:journals/corr/abs-2012-10289} is a benchmark dataset for explainable hate speech detection, sourced from social media platforms. Each post in the dataset is annotated from three perspectives: a three-class classification (hate, offensive, or normal), the targeted community, and rationales highlighting the specific text spans that justify the annotations. \\
\noindent\textbf{MedNLI.}\quad
MedNLI \citep{romanov-shivade-2018-lessons} is a specialized natural language inference (NLI) dataset tailored for the clinical domain. It comprises sentence pairs extracted from the Past Medical History sections of MIMIC-III clinical notes, annotated by physicians to determine whether a given hypothesis can be inferred from a premise.
\\
\noindent\textbf{MedMCQA.}\quad
MedMCQA \citep{pal2022medmcqalargescalemultisubject} is a comprehensive multiple-choice question answering dataset designed to emulate real-world medical entrance examinations. Each question is accompanied by multiple answer options and detailed explanations.
\\
For the train:validation:test size split and the number of samples in each dataset used for the evaluation, see Table \ref{tab:dataset_sizes}.

\begin{table}[h]
    \centering
    \scalebox{0.94}{
      \begin{tabular}{lccc}
          \hline
          \hline
          \textbf{Dataset} & \textbf{Train Size} & \textbf{Val. Size} & \textbf{Test Size} \\
          \hline
          \hline
          SST-2 & 67.3k & 0.8k & 1.8k \\
          MRPC & 3.6k & 0.4k & 1.7k \\
          QQP & 363.8k & 40.4k & 390.9k \\
          MNLI\textsubscript{m} & 392.7k & 9.8k & 9.8k \\
          MNLI\textsubscript{mm} & 392.7k & 9.8k & 9.8k \\
          QNLI & 104.7k & 5.4k & 5.4k \\
          RTE & 2.4k & 0.3k & 3.0k \\
          \hline
          AG News & 120.0k & 7.6k & -- \\
          IMDb & 25.0k & 25.0k & -- \\
          Yelp-5 & 650.0k & 50.0k & -- \\
          \hline
          MedNLI & 11.2k & 1.4k & 1.4k \\
          MedMCQA & 182.8k & 4.2k & 6.2k \\
          HateXplain & 16.0k & 2.0k & 2.0k \\    
          \hline
          \hline
      \end{tabular}
    }
    \caption{Dataset Sizes}
    \label{tab:dataset_sizes}
\end{table}

\subsection{Detailed Inference and Tuning Algorithm} \label{train_alg}
The algorithm \ref{alg:training_inference} outlines the explicit steps of SALSA's approach, covering both the training and inference flows.
 While using LLMs for classification at inference time is not a novel concept, steps 5 and 6 distinguish SALSA by showing how it leverages LLMs not just for auto-generation, but also for their underlying statistical properties - resulting in a richer and more informative output representation, and consequently, better performance.
During training, SALSA goes beyond the generic objective of predicting the correct next token for every position. Instead, it focuses specifically on the task-related tokens and updates the model weights based solely on the loss computed from these tokens, making the training process more efficient and better aligned with the classification objectives.
\begin{algorithm*}
    \centering
    \setlength{\columnwidth}{\textwidth}
    
    \caption{SALSA's \textcolor{blue}{Training} and Inference for Single-Task, Single-Label, Multi-Class Classification}
    \label{alg:training_inference}
    \begin{algorithmic}[1] 
    \Require $\text{instructions}, \text{answer template}, \text{answer's start}$ \Comment{Input parameters}
    \State \textbf{Definition:} Let $N$ be the vocabulary size.
    \For{each $s$ in \text{samples-to-classify}}
        \State $x\gets \text{wrap in the method's notation and tokenize}(s , input\_parameters)$
        \State $\text{logits} \gets \text{model's forward\_pass}(x)$
        \Comment{$\text{logits' size} = | \text{input} | \times N$}
        \State $y_{placeholder} \gets \text{logits}[\text{placeholder}]$
        \Comment{$y_{placeholder}\text{'s size} = N$}
        \State $y_{relevant} \gets y_{placeholder}[\text{categories}]$
        \Comment{$y_{relevant}$'s size =$|\text{categories}|$}
        \State $y_{\text{prob}} \gets \text{softmax}(y_{relevant})$
        
        \State \textcolor{blue}{${y_{true}} \gets \text{one\_hot}(\text{true\_label}, |\text{categories}|)$}
        \State \textcolor{blue}{$\text{loss} \gets \text{cross\_entropy}(y_{\text{prob}}, {y_{true}})$}
        \State \textcolor{blue}{$\text{model.backward\_pass}(\text{loss})$}
        \State \textcolor{blue}{$\text{update\_parameters}()$}
    
        \State report $\arg\max(y_{\text{prob}})$
    \EndFor
    \end{algorithmic}
    \smallskip
    \small{\textcolor{blue}{Note: The blue-colored lines correspond to training-specific steps.}}
\end{algorithm*}

\subsection{Training Details} \label{training_details}

The base model was Meta's Instruct LLama 3.3 70b (\href{https://www.llama.com/llama3_3/use-policy}{Meta's license}). It was tuned for a total of 6 epochs, and gradient accumulation steps set to 50 with batch size 1 to effectively handle large batch sizes in limited memory environment. To ensure reproducibility, a fixed random seed was used throughout the experiments.

LoRA\cite{hu2021lora} was used for fine-tuning, the rank was set to 8, the alpha parameter to 16, and a dropout rate of 0.05. It is 103M trainable parameters.

Optimization was carried out using the Adam optimizer \cite{kingma2014adam} with default parameter settings, where beta1=0.9, beta2=0.999, and epsilon=1E-8. A linear learning rate scheduler was employed, incorporating 100 warmup steps to progressively increase the learning rate at the beginning of training to 1E-4. After warmup the learning rate was reduced linearly to 0. For each experiment, the best-performing validation epoch was identified, and the experiment was repeated five times with different data shuffling seeds to ensure robustness of results.

Empirical observations revealed that optimal validation performance was typically achieved within the first 2 to 3 epochs. Training beyond this point, particularly when each sample was seen more than three times, often resulted in overfitting for small size datasets. The hardware used for this work was the Nvidia DGX system with eight H100 80GB GPU blades, and each model training run lasted between 1 and 36 hours. In this work, no hyperparameter optimization was conducted.

\subsection{Prompt Construction Example} \label{prompt_exm}

Figure~\ref{fig:prompt_construction} shows a sample prompt compiled from the RTE dataset. The prompt follows the default chat template of Instruct LLaMA 3.3, beginning with a default system prompt, followed by a user prompt containing task-specific instructions and data, and ending with the assistant response template. The compiled prompt is tokenized and processed in a single forward pass through the LLM to produce a classification output.

\usetikzlibrary{decorations.pathreplacing,calc}

\begin{figure*}[htbp]

\begin{tikzpicture}

\node[anchor=north east, inner sep=0pt] (box) at (12,0) {%
\fbox{%
\begin{varwidth}{0.8\textwidth}
\ttfamily
\footnotesize
\noindent
\textless|begin\_of\_text|\textgreater\\
\textless|start\_header\_id|\textgreater system \textless|end\_header\_id|\textgreater\\
Cutting Knowledge Date: December 2023 Today Date: 26 Jul 2024 \textless|eot\_id|\textgreater\\
\textless|start\_header\_id|\textgreater user \textless|end\_header\_id|\textgreater\\
Given the premise: \\
\textless PRE\-MISE\textgreater~Mangla was summoned after Madhumita's sister Nidhi Shukla,\\
who was the first witness in the case. \textless/ PRE\-MISE\textgreater \\
and hypothesis: \\
\textless HYPOTHESIS\textgreater~Shukla is related to Mangla. \textless/ HYPOTHESIS\textgreater \\
Is the hypothesis entailed by the premise? \\
Provide answer in format: \textless ANSWER\textgreater\#Number\textless/ANSWER\textgreater \\
where the number is one of the following: \\
0 - entailment \\
1 - not entailment \\
\textless|eot\_id|\textgreater\\
\textless|start\_header\_id|\textgreater assistant \textless|end\_header\_id|\textgreater\\
\textless ANSWER\textgreater X \textless/ANSWER\textgreater\\
\textless|eot\_id|\textgreater
\end{varwidth}
}};

\path (box.north west) coordinate (NW);

\draw [decorate,decoration={brace,amplitude=4pt,mirror}, thick]
  ($(NW)+(-0.0,-0.7)$) -- ($(NW)+(-0.0,-1.2)$)
  node[midway,xshift=-1.5cm]{\textbf{System prompt}};

\draw [decorate,decoration={brace,amplitude=4pt,mirror}, thick]
  ($(NW)+(-0.0,-1.5)$) -- ($(NW)+(-0.0,-4.1)$)
  node[midway,xshift=-1.3cm]{\textbf{Task + input}};

\draw [decorate,decoration={brace,amplitude=4pt,mirror}, thick]
  ($(NW)+(-0.0,-4.3)$) -- ($(NW)+(-0.0,-4.9)$)
  node[midway,xshift=-1.5cm]{\textbf{Class mapping}};

\draw [decorate,decoration={brace,amplitude=4pt,mirror}, thick]
  ($(NW)+(-0.0,-5.7)$) -- ($(NW)+(-0.0,-6.1)$)
  node[midway,xshift=-1.6cm]{\begin{tabular}{l}\textbf{Answer format} \\ with masked token\end{tabular}};

\end{tikzpicture}

\caption{A compiled prompt from RTE dataset before applying a forward pass.}
\label{fig:prompt_construction}
\end{figure*}
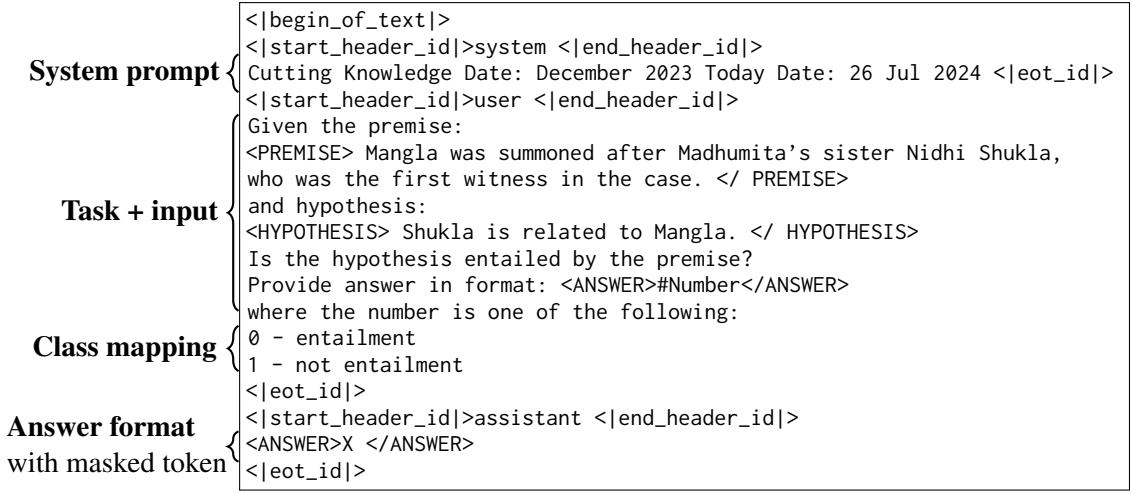

\subsection{Ablation Study on Label Mapping Strategies} \label{sensetivity}

We performed an ablation study on the RTE dataset to evaluate how different label mapping schemes affect classification performance. Specifically, we tested six mappings: numerical ('0', '1'), reverse numerical ('1', '0'), alphabetical ('A', 'B'), reverse alphabetical ('B', 'A'), semantic ('Y', 'N'), and reverse semantic ('N', 'Y').

The inclusion of “Y” and “N” label tokens was motivated by their implicit alignment with natural language concepts of affirmation and negation (“Yes”/“No”). We hypothesized that when the token aligns semantically with the correct label—e.g., “Y” for entailment—the model may perform better in a zero-shot setting. Conversely, using misleading or contradictory mappings, such as assigning “N” to entailment, could degrade performance due to interference with prior token associations.

Table~\ref{tab:mapping-ablation} summarizes the average accuracy (mean ± standard deviation over 5 runs) for each mapping strategy, evaluated in both zero-shot and fine-tuned conditions.

\begin{table}[htbp!]
    \centering
    \scalebox{0.95}{\begin{tabular}{lcc}
    \hline
    \hline
    \textbf{Mapping Strategy} & \textbf{Zero-Shot} & \textbf{Finetuned}\\
    \hline
    Numerical & 90.6 & 95.1 ± 0.4 \\
    Reverse Numerical & 89.1 & 94.4 ± 0.2 \\
    Alphabetical & 91.3 & 94.5 ± 0.3 \\
    Reverse Alphabetical & 90.2 & 94.2 ± 0.3 \\
    Semantic & 85.5 & 95.4 ± 0.7 \\
    Reverse Semantic & 44.7 & 94.0 ± 0.2 \\
    \textbf{Mean} & \textbf{81.9 ± 18} & \textbf{94.9 ± 0.5} \\
    \hline
    \hline    
    \end{tabular}}
\caption{Ablation study on different label mapping strategies for the RTE dataset. Accuracy is reported as mean ± std over 5 runs.}
\label{tab:mapping-ablation}
\end{table}

In the zero-shot setting, we observe substantial variation in performance across mappings. Reversing the labels (“N/Y”) led to the poorest performance, suggesting that mismatches between token semantics and label intent can confuse the model. After finetuning the differences between mappings diminish considerably, with all variants converging to similar accuracy levels. These findings confirm that the finetuning process effectively suppresses sensitivity to the mapping choices and enables the model to adapt even in the presence of initially misleading token associations.

\subsection{Discrete to Continuous Extension} \label{discrete_to_cont}

For tasks involving continuous value estimation, such as the STS-B benchmark \citep{cer-etal-2017-semeval} from the GLUE, we adapt our method to produce scalar outputs through a discretization-based approach.

We represent the predicted score as the expected value over a fixed set of discrete scalar values. Each value corresponds to a predefined class token and is associated with a probability derived from the model’s output distribution. Formally, let \(\mathcal{S} = \{s_1, s_2, \dots, s_n\}\) denote the set of discrete values (e.g., \([0.0, 0.2, \dots, 5.0]\)), and let \(P(s_i \mid x)\) be the probability assigned to state \(s_i\) given input \(x\). The model prediction \(\hat{y}\) is computed as:
\begin{equation}
 \label{eq:mean_value}
 \hat{y} = \sum_{i=1}^{n} P(s_i \mid x) \cdot s_i
\end{equation}

During training on STS-B, we perform the inverse operation. Given a scalar label \(y \in [0, 5]\), we identify the two discrete values \(s_i\) and \(s_{i+1}\) such that \(s_i \leq y \leq s_{i+1}\), and assign probabilities:
\begin{align}
\label{eq:inverse_representation}
    y &= \alpha \cdot s_i + (1 - \alpha) \cdot s_{i+1}, \\
    P(s_i \mid x) &= \alpha, \\
    P(s_{i+1} \mid x) &= 1 - \alpha
\end{align}

This construction ensures that the expected value of the predicted distribution matches the ground truth during supervision, while keeping the label space discrete and aligned with our logit-based framework.

Our method achieves a Pearson/Spearman correlation of \textbf{93.8/93.6} on the STS-B test set, compared to Turing v5's \textbf{93.7/93.3}, representing a new state-of-the-art result.

\subsection{Possible extention} \label{future}
\begin{figure*}[h]
  \centering
  \includegraphics[width=\textwidth]{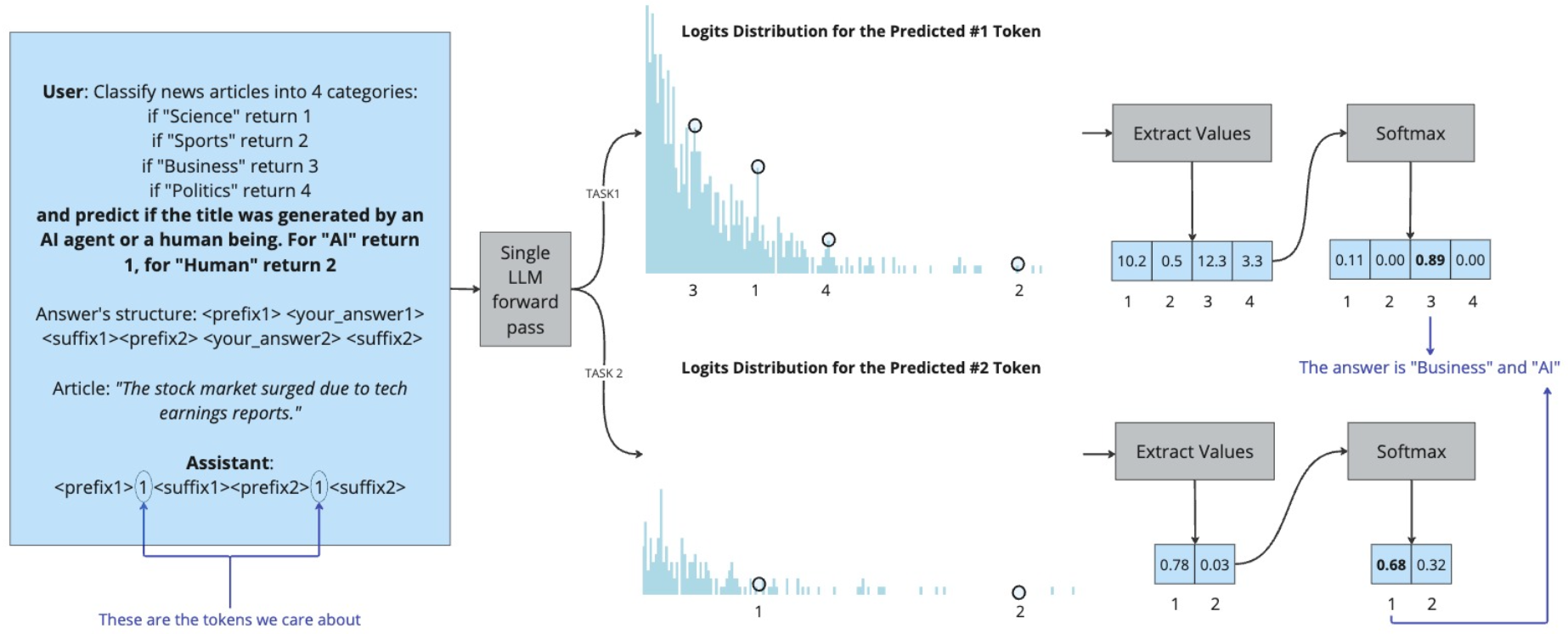}
  \caption{SALSA two-token classification pipeline: the LLM’s logits are used in a single pass to predict both the article’s topic (1–4) and its source (AI=1 or Human=2).}
  \label{fig:multi_class_method_diagram}
\end{figure*}

SALSA's framework can be naturally extended to more complex scenarios. For multi-label classification, one can replace the softmax layer with a sigmoid function and apply a probability threshold to select all relevant classes. For multi-task classification, a prompt with placeholders for each task enables the extraction of separate logits distributions, allowing simultaneous classification across multiple tasks in a single forward pass (see Figure \ref{fig:multi_class_method_diagram}).

\end{document}